\documentclass[a4paper]{article}
\usepackage[utf8]{inputenc}
\usepackage{natbib}
\usepackage{graphicx}
\usepackage{amsmath}
\usepackage[parfill]{parskip}
\usepackage{hyperref}
\usepackage[nameinlink,noabbrev]{cleveref}
\usepackage{amssymb}
\usepackage[printonlyused, withpage, nolist,nohyperlinks]{acronym}
\usepackage[]{algorithm2e}
\usepackage{booktabs} 
\usepackage{newunicodechar} 
\newunicodechar{∞}{$\infty$}
\usepackage[table]{xcolor}
\usepackage{makecell}
\usepackage{caption}
\usepackage[margin=1in]{geometry}
\usepackage[]{authblk}
\usepackage[all]{nowidow}

\newlength\kwInLength
\newcommand\kwInMultiline[1]{%
  \settowidth\kwInLength{\KwIn{}}%
  \setlength\hangindent{\kwInLength}%
  \hspace*{\kwInLength}#1\\}

\mathchardef\mhyphen="2D

\graphicspath{ {images/} }

\title{Oversampling for Imbalanced Learning\\Based on K-Means and SMOTE}
\author[1,*]{Felix Last}
\author[1]{Georgios Douzas}
\author[1]{Fernando Bacao}
\affil[1]{\footnotesize NOVA Information Management School, Universidade Nova de Lisboa}
\affil[*]{Corresponding author: mail@felixlast.de}
\affil[ ]{}
\affil[ ]{Postal Address: NOVA Information Management School, Campus de Campolide, 1070-312 Lisboa, Portugal \protect\\ 
	Telephone: +351 21 382 8610}
\date{}

\begin{document}
\maketitle

\begin{abstract}
Learning from class-imbalanced data continues to be a common and challenging problem in supervised learning as standard classification algorithms are designed to handle balanced class distributions. While different strategies exist to tackle this problem, methods which generate artificial data to achieve a balanced class distribution are more versatile than modifications to the classification algorithm. Such techniques, called oversamplers, modify the training data, allowing any classifier to be used with class-imbalanced datasets. Many algorithms have been proposed for this task, but most are complex and tend to generate unnecessary noise. This work presents a simple and effective oversampling method based on k-means clustering and SMOTE oversampling, which avoids the generation of noise and effectively overcomes imbalances between and within classes. Empirical results of extensive experiments with 71 datasets show that training data oversampled with the proposed method improves classification results. Moreover, k-means SMOTE consistently outperforms other popular oversampling methods. An implementation is made available in the python programming language.
\end{abstract}

\begin{acronym}[TT]
	\acro{A-SUWO}{adaptive semi-unsupervised weighted oversampling}
	\acro{AUPRC}{area under the precision-recall curve}
	\acro{GBM}{gradient boosting machine}
	\acro{KNN}{k-nearest neighbors}	
	\acro{LR}{logistic regression}
	\acro{PR}{precision-recall}
	\acro{SMOTE}{synthetic minority over-sampling technique}
	\acro{SOMO}{self-organizing map oversampling}

	\acrodefplural{LR}[LR]{Logistic regression}
	\acrodefplural{PR}[PR]{Precision-recall}
\end{acronym}

\pagenumbering{arabic} 

\section{Introduction}
The class imbalance problem in machine learning describes classification tasks in which classes of data are not equally represented. In many real-world applications, the nature of the problem implies a sometimes heavy skew in the class distribution of a binary or multi-class classification problem. Such applications include fraud detection in banking, rare medical diagnoses, and oil spill recognition in satellite images, all of which naturally exhibit a minority class~\citep{Chawla.2002,Kotsiantis.2006,Kotsiantis.2007,Galar.2012}.

The predictive capability of classification algorithms is impaired by class imbalance. Many such algorithms aim at maximizing classification accuracy, a measure which is biased towards the majority class. A classifier can achieve high classification accuracy even when it does not predict a single minority class instance correctly. For example, a trivial classifier which scores all credit card transactions as legit will score a classification accuracy of 99.9\% assuming that 0.1\% of transactions are fraudulent; however in this case, all fraud cases remain undetected. In conclusion, by optimizing classification accuracy, most algorithms assume a balanced class distribution~\citep{Provost.2000,Kotsiantis.2007}.

Another inherent assumption of many classification algorithms is the uniformity of misclassification costs, which is rarely a characteristic of real-world problems. Typically in imbalanced datasets, misclassifying the minority class as the majority class has a higher cost associated with it than vice versa. An example of this is database marketing, where the cost of mailing to a non-respondent is much lower than the lost profit of not mailing to a respondent~\citep{Domingos.1999}.

Lastly, what is referred to as the ``small disjuncts problem'' is often encountered in imbalanced datasets~\citep{Galar.2012}. The problem refers to classification rules covering only a small number of training examples. The presence of only few samples make rule induction more susceptible to error~\citep{Holte.1989}. To illustrate the importance of discovering high quality rules for sparse areas of the input space, the example of credit card fraud detection is again considered. Assume that most fraudulent transactions are processed outside the card owner's home country. The remaining cases of fraud happen within the country, but show some different exceptional characteristic, such as a high amount, an unusual time or recurring charges. Each of these other characteristics applies to only a very small group of transactions, which by itself is often vanishingly small. However, adding up all these edge cases, they can make up a substantial portion of all fraudulent transactions. Therefore, it is important that classifiers pay adequate attention to small disjuncts~\citep{Holte.1989}.

Techniques aimed at improving classification in the presence of class imbalance can be divided into three broad categories\footnote{The three categories are not exhaustive and new categories have been introduced, such as the combination of each of these techniques with ensemble learning~\citep{Galar.2012}.}: algorithm level methods, data level methods, and cost-sensitive methods.

Solutions which modify the classification algorithm to cope with the imbalance are algorithm level techniques~\citep{Kotsiantis.2006,Galar.2012}. Such techniques include changing the decision threshold and training separate classifiers for each class~\citep{Kotsiantis.2006,Chawla.2004}. 

In contrast, cost-sensitive methods aim at providing classification algorithms with different misclassification costs for each class. This requires knowledge of misclassification costs, which are dataset-dependent and commonly unknown or difficult to quantify. Additionally, the algorithms must be capable of incorporating the misclassification cost of each class or instance into their optimization. Therefore, these methods are regarded as operating both on data and algorithm level~\citep{Galar.2012}.

Finally, data-level methods manipulate the training data, aiming to change the class distribution towards a more balanced one. Techniques in this category resample the data by removing cases of the majority classes (undersampling) or adding instances to the minority classes by means of duplication or generation of new samples (oversampling)~\citep{Kotsiantis.2006,Galar.2012}. Because undersampling removes data, such methods risk the loss of important concepts. Moreover, when the number of minority observations is small, undersampling to a balanced distribution yields an undersized dataset, which may in turn limit classifier performance. Oversampling, on the other hand, may encourage overfitting when observations are merely duplicated~\citep{Weiss.2007}. This problem can be avoided by adding genuinely new samples. One straightforward approach to this is \ac{SMOTE}, which interpolates existing samples to generate new instances. 

Data-level methods can be further discriminated into random and informed methods. Unlike random methods, which randomly choose samples to be removed or duplicated (e.g. Random Oversampling, \ac{SMOTE}), informed methods take into account the distribution of the samples~\citep{Chawla.2004}. This allows informed methods to direct their efforts to critical areas of the input space, for instance to sparse areas~\citep{Nickerson.2001}, safe areas~\citep{Bunkhumpornpat.2009}, or to areas close to the decision boundary~\citep{Han.2005}. Consequently, informed methods may avoid the generation of noise and can tackle imbalances within classes.

Unlike algorithm-level methods, which are bound to a specific classifier, and cost-sensitive methods, which are problem-specific and need to be implemented by the classifier, data-level methods can be universally applied and are therefore more versatile~\citep{Galar.2012}.

Many oversampling techniques have proven to be effective in real-world domains. \ac{SMOTE} is the most popular oversampling method that was proposed to improve random oversampling. There are multiple variations of \ac{SMOTE} which aim to combat the original algorithm's weaknesses. Yet, many of these approaches are either very complex or alleviate only one of \ac{SMOTE}'s shortcomings. Additionally, few of them are readily available in a unified software framework used by practitioners.

This paper suggests the combination of the k-means clustering algorithm in combination with \ac{SMOTE} to combat some of other oversampler's shortcomings with a simple-to-use technique. The use of clustering enables the proposed oversampler to identify and target areas of the input space where the generation of artificial data is most effective. The method aims at eliminating both between-class imbalances and within-class imbalances while at the same time avoiding the generation of noisy samples. Its appeal is the widespread availability of both underlying algorithms as well the effectiveness of the method itself.

While the proposed method could easily be extended to cope with multi-class problems, the focus of this work is placed on binary classification tasks. When working with more than two imbalanced classes, different aspects of classification, as well as evaluation, must be considered, which is discussed in detail by \citet{Fernandez.2013}.

The remainder of this work is organized as follows. In \cref{sec:related-work}, related work is summarized and currently available oversampling methods are introduced. Special attention is paid to oversamplers which - like the proposed method - employ a clustering procedure. \Cref{sec:proposed-method} explains in detail how the proposed oversampling technique works and which hyperparameters need to be tuned. It is shown that both \ac{SMOTE} and random oversampling are limit cases of the algorithm and how they can be achieved. In \cref{sec:research-methodology}, a framework aimed at evaluating the performance of the proposed method in comparison with other oversampling techniques is established. The experimental results are shown in \cref{sec:experimental-results}, which is followed by \cref{sec:conclusion} presenting the conclusions.

\section{Related Work}
\label{sec:related-work}
Methods to cope with class imbalance either alter the classification algorithm itself, incorporate misclassification costs of the different classes into the classification process, or modify the data used to train the classifier. Resampling the training data can be done by removing majority class samples (undersampling) or by inflating the minority class (oversampling). Oversampling techniques either duplicate existing observations or generate artificial data. Such methods may work uninformed, randomly choosing samples which are replicated or used as a basis for data generation, or informed, directing their efforts to areas where oversampling is deemed most effective. Among informed oversamplers, clustering procedures are sometimes applied to determine suitable areas for the generation of synthetic samples.

Random oversampling randomly duplicates minority class instances until the desired class distribution is reached. Due to its simplicity and ease of implementation, it is likely to be the method that is most frequently used among practitioners. However, since samples are merely replicated, classifiers trained on randomly oversampled data are likely to suffer from overfitting\footnote{Overfitting occurs when a model does not conform with the principle of parsimony. A flexible model with more parameters than required is predisposed to fit individual observations rather than the overall distribution, typically impairing the model's ability to predict unseen data~\citep{Hawkins.2004}. In random oversampling, overfitting may occur when classifiers construct rules which seemingly cover multiple observations, while in fact they only cover many replicas of the same observation~\citep{Batista.2004}.}~\citep{Batista.2004,Chawla.2004}.

\begin{figure}[ht]
    \centering
	\includegraphics[width=1\textwidth]{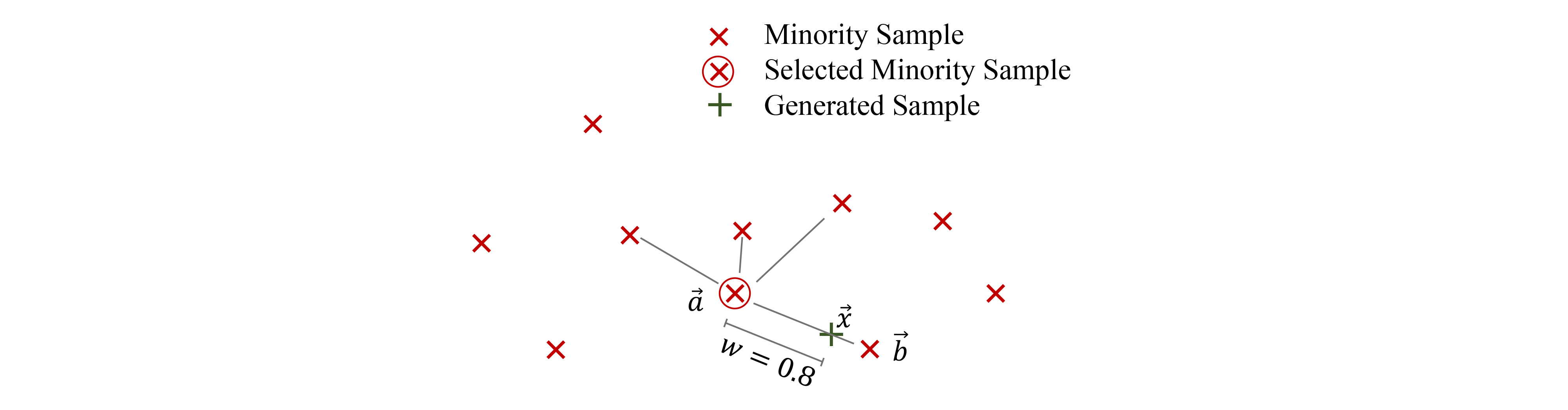}
	\caption{\acs{SMOTE} linearly interpolates a randomly selected minority sample and one of its $k=4$ nearest neighbors}
	\label{fig:smote-illustration}
\end{figure}

In 2002, \citet{Chawla.2002} suggested the \ac{SMOTE} algorithm, which avoids the risk of overfitting faced by random oversampling. Instead of merely replicating existing observations, the technique generates artificial samples. As shown in  \cref{fig:smote-illustration}, this is achieved by linearly interpolating a randomly selected minority observation and one of its neighboring minority observations. More precisely, \ac{SMOTE} executes three steps to generate a synthetic sample. Firstly, it chooses a random minority observation $\vec{a}$. Among its $k$ nearest minority class neighbors, instance $\vec{b}$ is selected. Finally, a new sample $\vec{x}$ is created by randomly interpolating the two samples: $\vec{x} = \vec{a} + w \times (\vec{b} - \vec{a})$, where $w$ is a random weight in $[0,1]$.

\begin{figure}[ht]
    \centering
	\includegraphics[width=1\textwidth]{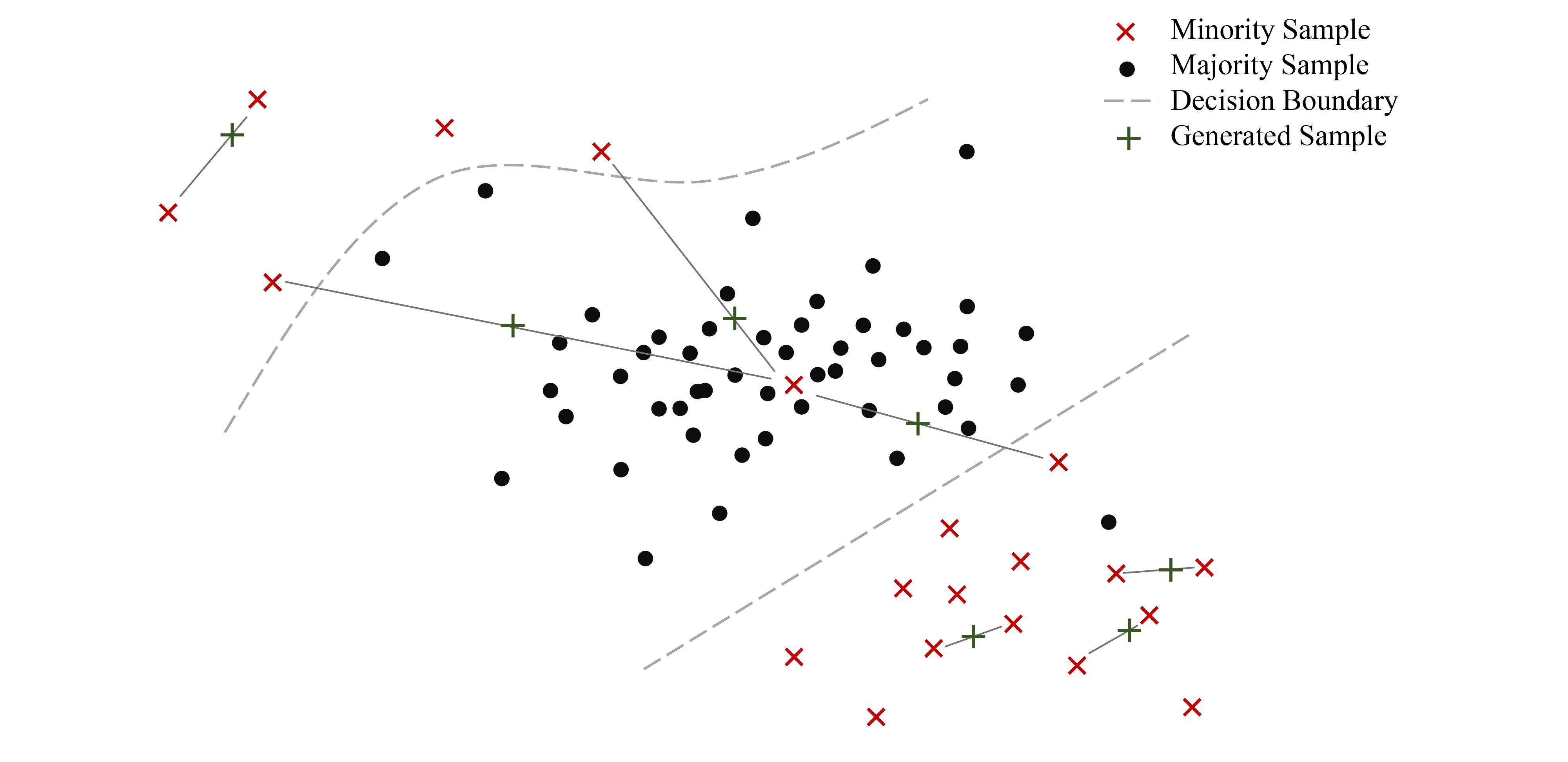}
	\caption{Behavior of \acs{SMOTE} in the presence of noise and within-class imbalance} 
	\label{fig:smote-noisy-samples}
\end{figure}

However, the algorithm has some weaknesses dealing with imbalance and noise as illustrated in \cref{fig:smote-noisy-samples}. One such drawback stems from the fact that \ac{SMOTE} randomly chooses a minority instance to oversample with uniform probability. While this allows the method to effectively combat between-class imbalance, the issues of within-class imbalance and small disjuncts are ignored. Input areas counting many minority samples have a high probability of being inflated further, while sparsely populated minority areas are likely to remain sparse~\citep{Prati.2004}.

Another major concern is that \ac{SMOTE} may further amplify noise present in the data. This is likely to happen when linearly interpolating a noisy minority sample, which is located among majority class instances, and its nearest minority neighbor. The method is susceptible to noise generation because it doesn't distinguish overlapping class regions from so-called safe areas~\citep{Bunkhumpornpat.2009}. 

Finally, the algorithm does not specifically enforce the decision boundary. Instances far from the class border are oversampled with the same probability as those close to the boundary. It has been argued that classifiers could benefit from the generation of samples closer to the class border~\citep{Han.2005}.

Despite its weaknesses, \ac{SMOTE} has been widely adopted by researchers and practitioners, likely due to its simplicity and added value with respect to random oversampling. Numerous extensions of the technique have been developed, which aim to eliminate its disadvantages. Such extensions typically address one of the original method's weaknesses. They may be divided according to their claimed goal into algorithms which aim to emphasize certain minority class regions, intend to combat within-class imbalance, or attempt to avoid the generation of noise.

Focussing its attention on the decision boundary, borderline-\ac{SMOTE}1 belongs to the category of methods emphasizing class regions. It is the only algorithm discussed here which does not employ a clustering procedure and is included due to its popularity. The technique replaces \ac{SMOTE}'s random selection of observations with a targeted selection of instances close to the class border. The label of a sample's $k$ nearest neighbors is used to determine whether it is discarded as noise, selected for its presumed proximity to the class border, or ruled out because it is far from the boundary. Borderline-\ac{SMOTE}2 extends this method to allow interpolation of a minority instance and one of its \textit{majority} class neighbors, setting the interpolation weight to less than 0.5 so as to place the generated sample closer to the minority sample~\citep{Han.2005}.

Cluster-SMOTE, another method in the category of techniques emphasizing certain class regions, uses k-means to cluster the minority class before applying \ac{SMOTE} within the found clusters. The stated goal of this method is to boost class regions by creating samples within naturally occurring clusters of the minority class. It is not specified how many instances are generated in each cluster, nor how the optimal number of clusters can be determined~\citep{Cieslak.2006b}. While the method may alleviate the problem of between-class imbalance, it does not help to eliminate small disjuncts.

Belonging to the same category, \ac{A-SUWO} introduced by \citet{Nekooeimehr.2016}, is a rather complex technique which applies clustering to improve the quality of oversampling. The approach is based on hierarchical clustering and aims at oversampling hard-to-learn instances close to the decision boundary.

Among techniques which aim to reduce within-class imbalance at the same time as between-class imbalance is cluster-based oversampling. The algorithm clusters the entire input space using k-means. Random oversampling is then applied within clusters so that: a) all majority clusters, b) all minority clusters, and c) the majority and minority classes are of the same size~\citep{Jo.2004}. By replicating observations instead of generating new ones, this technique may encourage overfitting.

With a bi-directional sampling approach, \citet{Song.2016} combine undersampling the majority class with oversampling the minority class. K-means clustering is applied separately within each class with the goal of achieving within- and between-class balance. For clustering the majority class, the number of clusters is set to the desired number of samples (equal to the geometric mean of instances per class). The class is undersampled by retaining only the nearest neighbor of each cluster centroid. The minority class is clustered into two partitions. Subsequently, \ac{SMOTE} is applied in the smaller cluster. A number of iterations of clustering and \ac{SMOTE} are performed until both classes are of equal size. It is unclear how many samples are added at each iteration. Since the method clusters both classes separately, it is blind to overlapping class borders and may contribute to noise generation.

The \ac{SOMO} algorithm transforms the input data into a two-dimensional space using a self-organizing map, where safe and effective areas are identified for data generation. \ac{SMOTE} is then applied within clusters found in the lower dimensional space, as well as between neighboring clusters in order to correct within- and between-class imbalances~\citep{Douzas.2017}.

Aiming to avoid noise generation, a clustering-based approach called CURE-SMOTE uses the hierarchical clustering algorithm CURE to clear the data of outliers before applying \ac{SMOTE}. The rationale behind this method is that because \ac{SMOTE} would amplify existing noise, the data should be cleared of noisy observations prior to oversampling~\citep{Ma.2017}. While noise generation is avoided, possible imbalances within the minority class are ignored.

Finally, \citet{Santos.2015} cluster the entire input space with k-means. Clusters with few representatives are chosen to be oversampled using SMOTE. The algorithm is different from most oversampling methods in that \ac{SMOTE} is applied regardless of the class label. The class label of the generated sample is copied from the nearest of the two parents. The algorithm thereby targets dataset imbalance, rather than imbalances between or within classes and cannot be used to solve class imbalance.

In summary, there has been a lot of recent research aimed at the improvement of imbalanced dataset resampling. Some proposed methods employ clustering techniques before applying random oversampling or \ac{SMOTE}. While most of them manage to combat some weaknesses of existing oversampling algorithms, none have been shown to avoid noise generation and alleviate imbalances both between and within classes at the same time. Additionally, many techniques achieve their respective improvements at the cost of high complexity, making the techniques difficult to implement and use.

\section{Proposed Method}
\label{sec:proposed-method}
The method proposed in this work employs the simple and popular k-means clustering algorithm in conjunction with \ac{SMOTE} oversampling in order to rebalance skewed datasets. It manages to avoid the generation of noise by oversampling only in safe areas. Moreover, its focus is placed on both between-class imbalance and within-class imbalance, combating the small disjuncts problem by inflating sparse minority areas. The method is easily implemented due to its simplicity and the widespread availability of both k-means and \ac{SMOTE}. It is uniquely different from related methods not only due to its low complexity but also because of its effective approach to distributing synthetic samples based on cluster density.

	\subsection{Algorithm}
\begin{figure}[ht]
	\centering
	\includegraphics[width=1\textwidth]{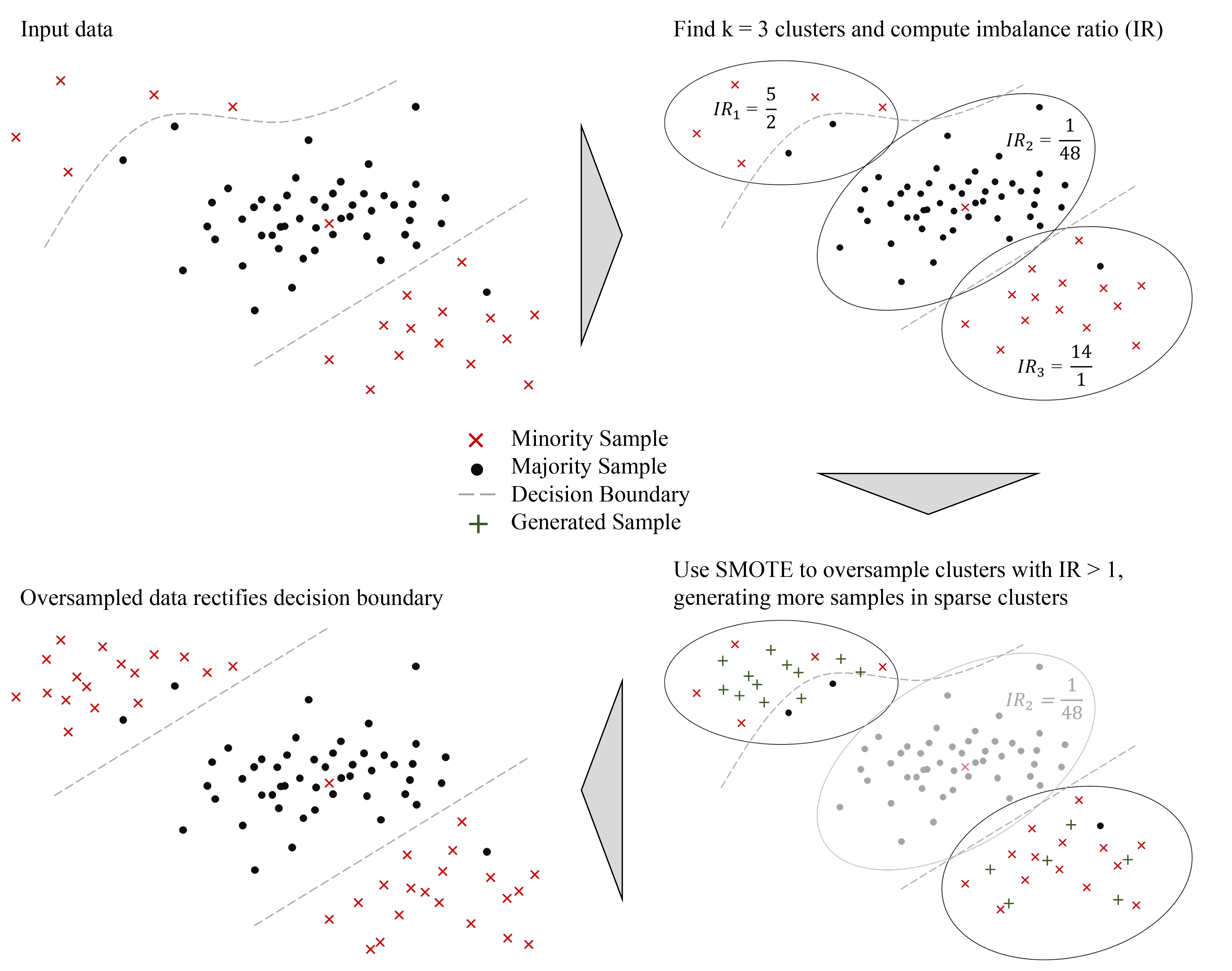}
	\caption[K-means \acs{SMOTE} oversamples safe areas and combats within-class imbalance]{K-means \ac{SMOTE} oversamples safe areas and combats within-class imbalance\protect\footnotemark}
    \label{fig:kmeans-smote-illustration}
	\end{figure}

	K-means \ac{SMOTE} consists of three steps: clustering, filtering, and oversampling. In the clustering step, the input space is clustered into $k$ groups using k-means clustering. The filtering step selects clusters for oversampling, retaining those with a high proportion of minority class samples. It then distributes the number of synthetic samples to generate, assigning more samples to clusters where minority samples are sparsely distributed. Finally, in the oversampling step, \ac{SMOTE} is applied in each selected cluster to achieve the target ratio of minority and majority instances. The algorithm is illustrated in \cref{fig:kmeans-smote-illustration}.

	The k-means algorithm is a popular iterative method of finding naturally occurring groups in data which can be represented in a Euclidean space. It works by iteratively repeating two instructions: First, assign each observation to the nearest of $k$ cluster centroids. Second, update the position of the centroids so that they are centered between the observations assigned to them. The algorithm converges when no more observations are reassigned. It is guaranteed to converge to a typically local optimum in a finite number of iterations~\citep{MacQueen.1967}. For large datasets where k-means may be slow to converge, more efficient implementations could be used for the clustering step of the k-means \ac{SMOTE}, such as mini-batch k-means as proposed by~\citep{Sculley.2010}. All hyperparameters of k-means are also hyperparameters of the proposed algorithm, most notably $k$, the number of clusters. Finding an appropriate value for $k$ is essential for the effectiveness of k-means \ac{SMOTE} as it influences how many minority clusters, if any, can be found in the filter step.

	\footnotetext{The rectification of the decision boundary in the lower left of the image is desired because the two majority samples are considered outliers. The classifier is thus able to induce a simpler rule, which is less error prone.}

	Following the clustering step, the filter step chooses clusters to be oversampled and determines how many samples are to be generated in each cluster. The motivation of this step is to oversample only clusters dominated by the minority class, as applying \ac{SMOTE} inside minority areas is less susceptible to noise generation. Moreover, the goal is to achieve a balanced distribution of samples \textit{within} the minority class. Therefore, the filter step allocates more generated samples to sparse minority clusters than to dense ones.
	
	The selection of clusters for oversampling is based on each cluster's proportion of minority and majority instances. By default, any cluster made up of at least 50 \% minority samples is selected for oversampling. This behavior can be tuned by adjusting the imbalance ratio threshold (or $irt$), a hyperparameter of k-means \ac{SMOTE} which defaults to $1$. The imbalance ratio of a cluster $c$ is defined as $\frac{majorityCount(c) + 1}{minorityCount(c) + 1}$. When the imbalance ratio threshold is increased, cluster choice is more selective and a higher proportion of minority instances is required for a cluster to be selected. On the other hand, lowering the threshold loosens the selection criterion, allowing clusters with a higher majority proportion to be chosen.

	To determine the distribution of samples to be generated, filtered clusters are assigned sampling weights between zero and one. A high sampling weight corresponds to a low density of minority samples and yields more generated samples. To achieve this, the sampling weight depends on how dense a single cluster is compared to how dense all selected clusters are on average. Note that when measuring a cluster's density, only the distances among minority instances are considered. The computation of the sampling weight may be expressed by means of five sub-computations:
	\begin{enumerate}
		\item For each filtered cluster $f$, calculate the Euclidean distance matrix, ignoring majority samples.
		\item Compute the mean distance within each cluster by summing all non-diagonal elements of the distance matrix, then dividing by the number non-diagonal elements.
		\item To obtain a measure of density, divide each cluster's number of minority instances by its average minority distance raised to the power of the number of features $m$: $density(f) = \frac{minorityCount(f)}{averageMinorityDistance(f)^{m}}$.
		\item Invert the density measure as to get a measure of sparsity, i.e. $sparsity(f) = \frac{1}{density(f)}$.
		\item The sampling weight of each cluster is defined as the cluster's sparsity factor divided by the sum of all clusters' sparsity factors.
	\end{enumerate}
	 Consequently, the sum of all sampling weights is one. Due to this property, the sampling weight of a cluster can be multiplied by the overall number of samples to be generated to determine the number of samples to be generated in that cluster.

	In the oversampling step of the algorithm, each filtered cluster is oversampled using \ac{SMOTE}. For each cluster, the oversampling procedure is given all points of the cluster along with the instruction to generate $\Vert samplingWeight(f) \times n \Vert$ samples, where n is the overall number of samples to be generated. Per synthetic sample to generate, \ac{SMOTE} chooses a random minority observation $\vec{a}$ within the cluster, finds a random neighboring minority instance $\vec{b}$ of that point and determines a new sample $\vec{x}$ by randomly interpolating $\vec{a}$ and $\vec{b}$. In geometric terms, the new point $\vec{x}$ is thus placed somewhere along a straight line from $\vec{a}$ to $\vec{b}$. The process is repeated until the number of samples to be generated is reached. 
	
	\ac{SMOTE}'s hyperparameter k nearest neighbors, or $knn$, constitutes among how many neighboring minority samples of $\vec{a}$ the point $\vec{b}$ is randomly selected. This hyperparameter is also used by k-means \ac{SMOTE}. Depending on the specific implementation of \ac{SMOTE}, the value of $knn$ may have to be adjusted downward when a cluster has fewer than $knn + 1$ minority samples. Once each filtered cluster has been oversampled, all generated samples are returned and the oversampling process is completed.

	The proposed method is distinct from related techniques in that it clusters the entire dataset regardless of the class label. An unsupervised approach enables the discovery of overlapping class regions and may aid the avoidance of oversampling in unsafe areas. This is in contrast to cluster-SMOTE, where only minority class instances are clustered~\citep{Cieslak.2006b} and to the aforementioned combination of oversampling and undersampling where both classes are clustered separately~\citep{Song.2016}. Another distinguishing feature is the unique approach to the distribution of generated samples across clusters: sparse minority clusters yield more samples than dense ones. The previously presented method cluster-based oversampling, on the other hand, distributes samples based on cluster size~\citep{Jo.2004}. Since k-means may find clusters of varying density, but typically of the same size~\citep{MacQueen.1967}, distributing samples according to cluster density can be assumed to be an effective way to combat within-class imbalance. Lastly, the use of \ac{SMOTE} circumvents the problem of overfitting, which random oversampling has been shown to encourage.

	\begin{algorithm}[htbp]
	\DontPrintSemicolon
	\KwIn{$X$ (matrix of observations)}
	\kwInMultiline{$y$ (target vector)}
	\kwInMultiline{$n$ (number of samples to be generated)}
	\kwInMultiline{$k$ (number of clusters to be found by k-means)}
	\kwInMultiline{$irt$ (imbalance ratio threshold)}
	\kwInMultiline{$knn$ (number of nearest neighbors considered by \ac{SMOTE})}
	\kwInMultiline{$de$ (exponent used for computation of density; defaults to the number of features in $X$)}
	\Begin{
		\tcp{Step 1: Cluster the input space and filter clusters with more minority instances than majority instances.}
		$clusters \gets kmeans(X)$\;
		$filteredClusters \gets \varnothing$\;
		\For{$c \in clusters $}{
			$imbalanceRatio \gets \frac{majorityCount(c) + 1}{minorityCount(c) + 1}$\;
			\If{$imbalanceRatio < irt$}{
				$filteredClusters \gets filteredClusters \cup \{ c \}$\;
			}
		}
		
		\;
		\tcp{Step 2: For each filtered cluster, compute the sampling weight based on its minority density.}
		\For{$f \in filteredClusters $}{
			$averageMinorityDistance(f) \gets mean( euclideanDistances(f) )$\;
			$densityFactor(f) \gets \frac{minorityCount(f)}{averageMinorityDistance(f)^{de}}$\;
			$sparsityFactor(f) \gets \frac{1}{densityFactor(f)}$
		}
		$sparsitySum \gets \sum_{f \in filteredClusters}^{} sparsityFactor(f)$\;
		$samplingWeight(f) \gets \frac{sparsityFactor(f)}{sparsitySum}$\;

		\;
		\tcp{Step 3: Oversample each filtered cluster using \ac{SMOTE}. The number of samples to be generated is computed using the sampling weight.}
		$generatedSamples \gets \varnothing$\;
		\For{$f \in filteredClusters $}{
			$numberOfSamples \gets \Vert n \times samplingWeight(f) \Vert$\;
			$generatedSamples \gets generatedSamples \cup \{ SMOTE(f, numberOfSamples, knn \}$\;
		}
		\Return{$generatedSamples$}
	}
	\caption{Proposed method based on k-means and \ac{SMOTE}}
	\end{algorithm}

	\subsection{Limit Cases}
	In the following, it is shown that \ac{SMOTE} and random oversampling can be regarded as limit cases of the more general method proposed in this work. In k-means \ac{SMOTE}, the input space is clustered using k-means. Subsequently, some clusters are selected and then oversampled using \ac{SMOTE}. Considering the case where the number of clusters $k$ is equal to 1, all observations are grouped in one cluster. For this only cluster to be selected as a minority cluster, the imbalance ratio threshold needs to be set so that the imbalance ratio of the training data is met. For example, in a dataset with 100 minority observations and 10,000 majority observations, the imbalance ratio threshold must be greater than or equal to $\frac{10,000 + 1}{100 + 1} \approx 99.02$. The single cluster is then selected and oversampled using \ac{SMOTE}; since the cluster contains all observations, this is equivalent to simply oversampling the original dataset with \ac{SMOTE}. Instead of setting the imbalance ratio threshold to the exact imbalance ratio of the dataset, it can simply be set to positive infinity.

	If \ac{SMOTE} did not interpolate two different points to generate a new sample but performed the random interpolation of one and the same point, the result would be a copy of the original point. This behavior could be achieved by setting the parameter ``k nearest neighbors'' of \ac{SMOTE} to zero if the concrete implementation supports this behavior. As such, random oversampling may be regarded as a specific case of \ac{SMOTE}.

	This property of k-means \ac{SMOTE} is of very practical value to its users: since it contains both \ac{SMOTE} and random oversampling, a search of optimal hyperparameters could include the configurations for those methods. As a result, while a better parametrization may be found, the proposed method will perform at least as well as the better of both oversamplers. In other words, \ac{SMOTE} and random oversampling are fallbacks contained in k-means \ac{SMOTE}, which can be resorted to when the proposed method does not produce any gain with other parametrizations. \Cref{tab:limit-case} summarizes the parameters which may be used to reproduce the behavior of both algorithms.
	\begin{table}[!htb]
		\centering
		\rowcolors{2}{gray!10}{white}	
		\begin{tabular}{lrrrrr}
		\toprule
							& $k$ & $irt$    & $knn$ \\
		\midrule
		\ac{SMOTE}          & 1 & $\infty$ &     \\
		Random Oversampling & 1 & $\infty$ & 0  \\
		\bottomrule
		\end{tabular}
		\caption{Limit case configurations}
		\label{tab:limit-case}
	\end{table}

\section{Research Methodology}
\label{sec:research-methodology}
The ultimate purpose of any resampling method is the improvement of classification results. In other words, a resampling technique is successful if the resampled data it produces improves the prediction quality of a given classifier. Therefore, the effectiveness of an oversampling method can only be assessed indirectly by evaluating a classifier trained on oversampled data. This proxy measure, i.e. the classifier performance, is only meaningful when compared with the performance of the same classification algorithm trained on data which has not been resampled. Multiple oversampling techniques can then be ranked by evaluating a classifier's performance with respect to each modified training set produced by the resamplers.

A general concern in classifier evaluation is the bias of evaluating predictions for previously seen data. Classifiers may perform well when making predictions for rows of data used during training, but poorly when classifying new data. This problem is also referred to as overfitting. Oversampling techniques have been observed to encourage overfitting, which is why this bias should be carefully avoided during their evaluation. A general approach is to split the available data into two or more subsets of which only one is used during training, and another is used to evaluate the classification. The latter is referred to as the holdout set, unknown data, or test dataset. 

Arbitrarily splitting the data into two sets, however, may introduce additional biases. One potential issue that arises is that the resulting training set may not contain certain observations, preventing the algorithm from learning important concepts. Cross-validation combats this issue by randomly splitting the data many times, each time training the classifier from scratch using one portion of the data before measuring its performance on the remaining share of data. After a number of repetitions, the classifier can be evaluated by aggregating the results obtained in each iteration. In k-fold cross-validation, a popular variant of cross-validation, $k$ iterations, called folds, are performed. During each fold, the test set is one of $k$ equally sized groups. Each group of observations is used exactly once as a holdout set. K-fold cross-validation can be repeated many times to avoid potential bias due to random grouping~\citep{Japkowicz.2013}.

While k-fold cross validation typically avoids the most important biases in classification tasks, it might distort the class distributions when randomly sampling from a class-imbalanced dataset. In the presence of extreme skews, there may even be iterations where the test set contains no instances of the minority class, in which case classifier evaluation would be ill-defined or potentially strongly biased. A simple and common approach to this problem is to use stratified cross-validation, where instead of sampling completely at random, the original class distribution is preserved in each fold~\citep{Japkowicz.2013}.

	\subsection{Metrics}
	Of the various assessment metrics traditionally used to evaluate classifier performance, not all are suitable when the class distribution is not uniform. However, there are metrics which have been employed or developed specifically to cope with imbalanced data.

\begin{figure}[ht]
	\centering
	\includegraphics[width=0.6\textwidth]{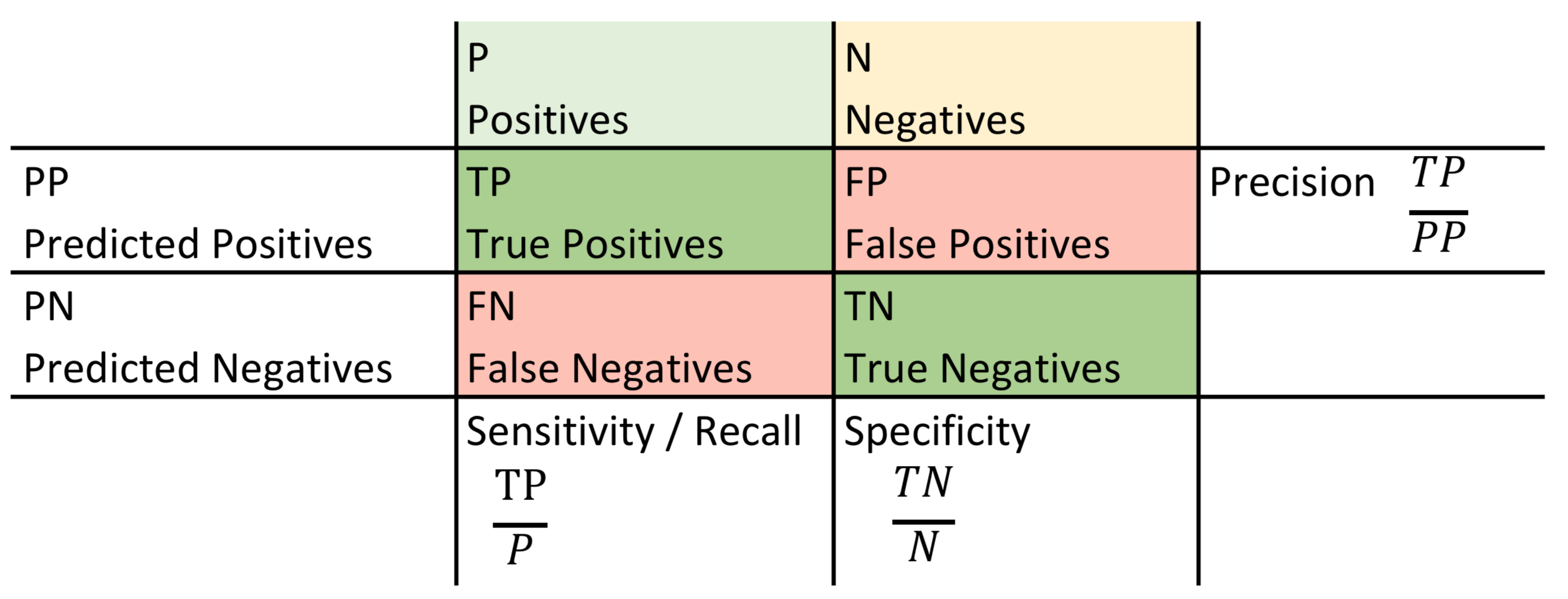}
	\caption{Confusion matrix}
    \label{fig:confusion-matrix}
	\end{figure}

	Classification assessment metrics compare the true class membership of each observation with the prediction of the classifier. To illustrate the alignment of predictions with the true distribution, a confusion matrix (\cref{fig:confusion-matrix}) can be constructed.	Possibly deriving from medical diagnoses, a positive observation is a rare case and belongs to the minority class. The majority class is considered negative~\citep{Japkowicz.2013}.

	When evaluating a single classifier in the context of a finite dataset with fixed imbalance ratio, the confusion matrix provides all necessary information to assess the classification quality. However, when comparing different classifiers or evaluating a single classifier in variable environments, the absolute values of the confusion matrix are non-trivial.

	The most common metrics for classification problems are accuracy and its inverse, error rate.
	\begin{equation}
	Accuracy = \frac{TP + TN}{P + N};\ Error Rate = 1 - Accuracy
	\end{equation}
	These metrics show a bias toward the majority class in imbalanced datasets. For example, a naive classifier which predicts all observations as negative would achieve 99\% accuracy in a dataset where only 1\% of instances are positive. While such high accuracy suggests an effective classifier, the metric obscures the fact that not a single minority instance was predicted correctly~\citep{He.2009}.

	 Sensitivity, also referred to as recall or true positive rate, explains the prediction accuracy among minority class instances. It answers the question ``How many minority instances were correctly classified as such?'' Specificity answers the same question for the majority class. Precision is the rate of correct predictions among all instances predicted to belong to the minority class. It indicates how many of the positive predictions are correct~\citep{He.2009}.

	 The F1-score, or F-measure, is the (often weighted) harmonic mean of precision and recall. In other terms, the indicator rates both the completeness and exactness of positive predictions~\citep{He.2009,Japkowicz.2013}.
	 
	\begin{equation}
	F1 = \frac{(1 + \alpha) \times (sensitivity \times precision)}{sensitivity + \alpha \times precision} 
	= \frac{(1 + \alpha) \times (\frac{TP}{P} \times \frac{TP}{PP})}{\frac{TP}{P} + \alpha \times \frac{TP}{PP} }
	\end{equation}

	The geometric mean score, also referred to as g-mean or g-measure, is defined as the geometric mean of sensitivity and specificity. The two components can be regarded as per-class accuracy. The g-measure aggregates both metrics into a single value in $[0,1]$, assigning equal importance to both~\citep{He.2009,Japkowicz.2013}.
	
	\begin{equation}
	g\mhyphen mean = \sqrt{sensitivity \times specificity} 
	= \sqrt{\frac{TP}{P} \times \frac{TN}{N}}
	\end{equation}

	\acp{PR} diagrams plot the precision of a classifier as a function of its minority accuracy. Classifiers outputting class membership confidences (i.e. continuous values in $[0,1]$) can be plotted as multiple points in discrete intervals, resulting in a \ac{PR} curve. Commonly, the \ac{AUPRC} is computed as a single numeric performance metric~\citep{He.2009,Japkowicz.2013}.
	
	The choice of metric depends to a great extent on the goal their user seeks to achieve. In certain practical tasks, one specific aspect of classification may be more important than another (e.g. in medical diagnoses, false negatives are much more critical than false positives). However, to determine a general ranking among oversamplers, no such focus should be placed. Therefore, the following unweighted metrics are chosen for the evaluation.
    \begin{itemize}
    	\item g-mean
        \item F1-score
		\item \ac{AUPRC}
	\end{itemize}
   
	\subsection{Oversamplers}
	The following list enumerates the oversamplers used as a benchmark for the evaluation of the proposed method, along with the set of hyperparameters used for each. The optimal imbalance ratio is not obvious and has been discussed by other researchers~\citep{Provost.2000,Estabrooks.2004}. This work aims at creating comparability among oversamplers; consequently, it is most important that oversamplers achieve the same imbalance ratio. Therefore, all oversampling methods were parametrized to generate as many instances as necessary so that minority and majority classes count the same number of samples.
	\begin{itemize}
		\item random oversampling
		\item SMOTE
		\begin{itemize}
			\item $knn \in \{3, 5, 20\}$
		\end{itemize}
		\item borderline-SMOTE1
		\begin{itemize}
			\item $knn \in \{3, 5, 20\}$
		\end{itemize}
		\item borderline-SMOTE2
		\begin{itemize}
			\item $knn \in \{3, 5, 20\}$
		\end{itemize}
		\item k-means SMOTE
		\begin{itemize}
			\item $k \in \{2, 20, 50, 100, 250, 500\}$
			\item $knn \in \{3, 5, 20, \infty\}$
			\item $irt \in \{1, \infty\}$
			\item $de \in \{0, 2, numberOfFeatures\}$
		\end{itemize}
		\item no oversampling
	\end{itemize}

	\subsection{Classifiers}
	For the evaluation of the various oversampling methods, several different classifiers are chosen to ensure that the results obtained can be generalized and are not constrained to the usage of a specific classifier. The choice of classifiers is further motivated by the number of hyperparameters: classification algorithms with few or no hyperparameters are less likely to bias results due to their specific configuration.

	\acp{LR} is a generalization of linear regression which can be used for binary classification. Fitting the model is an optimization problem which can be solved using simple optimizers which require no hyperparameters to be set~\citep{McCullagh.1984}. Consequently, results achieved by \ac{LR} are easily reproducible, while also constituting a baseline for more sophisticated approaches.

	Another classification algorithm referred to as \ac{KNN} assigns an observation to the class most of its nearest neighbors belong to. How many neighbors are considered is determined by the method's hyperparameter~$k$~\citep{Fix.1951}.

	Finally, gradient boosting over decision trees, or simply \ac{GBM}, is an ensemble technique used for classification. In the case of binary classification, one shallow decision tree is induced at each stage of the algorithm. Each tree is fitted to observations which could not be correctly classified by decision trees of previous stages. Predictions of \ac{GBM} are made by majority vote of all trees. In this way, the algorithm combines several simple models (referred to as weak learners) to create one effective classifier. The number of decision trees to generate, which in binary classification is equal to the number of stages, is a hyperparameter of the algorithm~\citep{Friedman.2001}. 

	As further explained in \cref{sec:experimental-framework}, various combinations of hyperparameters are tested for each classifier. All classifiers are used as implemented in the python library scikit-learn~\citep{Pedregosa.2011} with default parameters unless stated otherwise. The following list enumerates the classifiers used in this study along with a set of values for their respective hyperparameters.
	\begin{itemize}
		\item \ac{LR}
		\item \ac{KNN}
		\begin{itemize}
            \item $k \in \{3, 5, 8\}$
		\end{itemize}
		\item \ac{GBM}
		\begin{itemize}
            \item $numberOfTrees \in \{50, 100, 200\}$
		\end{itemize}
	\end{itemize}

	\subsection{Datasets}
    \label{sec:datasets}
	To evaluate k-means \ac{SMOTE}, 12 imbalanced datasets from the UCI Machine Learning Repository~\citep{Lichman.2013} are used. Those datasets containing more than two classes were binarized using a one-versus-rest approach, labeling the smallest class as the minority and merging all other samples into one class. In order to generate additional datasets with even higher imbalance ratios, each of the aforementioned datasets was randomly undersampled to generate up to six additional datasets. The imbalance ratio of each dataset was increased approximately by multiplication factors of 2, 4, 6, 10, 15 and 20, but only if a given factor did not reduce a dataset's total number of minority samples to less than eight. Furthermore, the python library scikit-learn~\citep{Pedregosa.2011} was used to generate ten variations of the artificial ``MADELON'' dataset, which poses a difficult binary classification problem~\citep{Guyon.2003}.

    \Cref{tab:datasets} lists the datasets used to evaluate the proposed method, along with important characteristics. The artificial datasets are referred to as simulated. Undersampled versions of the original datasets are omitted from the table. All datasets used in the study are made available at \url{https://github.com/felix-last/evaluate-kmeans-smote/releases/download/v0.0.1/uci_extended.tar.gz} for the purpose of reproducibility. 

	\begin{table}[!htb]
		\rowcolors{2}{gray!10}{white}	
		\begin{tabular}{lrrrrr}
			\toprule
				Dataset &  \# features &  \# instances &  \# minority instances &  \# majority instances &  imbalance ratio \\
			\midrule
				breast\_tissue &              9 &             106 &                       36 &                       70 &             1.94 \\
				ecoli &              7 &             336 &                       52 &                      284 &             5.46 \\
				glass &              9 &             214 &                       70 &                      144 &             2.06 \\
				haberman &              3 &             306 &                       81 &                      225 &             2.78 \\
				heart &             13 &             270 &                      120 &                      150 &             1.25 \\
					iris &              4 &             150 &                       50 &                      100 &             2.00 \\
				libra &             90 &             360 &                       72 &                      288 &             4.00 \\
				liver\_disorders &              6 &             345 &                      145 &                      200 &             1.38 \\
					pima &              8 &             768 &                      268 &                      500 &             1.87 \\
				segment &             16 &            2310 &                      330 &                     1980 &             6.00 \\
				simulated1 &             20 &            4000 &                       25 &                     3975 &           159.00 \\
				simulated2 &             20 &            4000 &                       23 &                     3977 &           172.91 \\
				simulated3 &             20 &            4000 &                       23 &                     3977 &           172.91 \\
				simulated4 &             20 &            4000 &                       26 &                     3974 &           152.85 \\
				simulated5 &             20 &            4000 &                       23 &                     3977 &           172.91 \\
				simulated6 &            200 &            3000 &                       20 &                     2980 &           149.00 \\
				simulated7 &            200 &            3000 &                       19 &                     2981 &           156.89 \\
				simulated8 &            200 &            3000 &                       15 &                     2985 &           199.00 \\
				simulated9 &            200 &            3000 &                       13 &                     2987 &           229.77 \\
				simulated10 &            200 &            3000 &                       22 &                     2978 &           135.36 \\
				vehicle &             18 &             846 &                      199 &                      647 &             3.25 \\
				wine &             13 &             178 &                       71 &                      107 &             1.51 \\
			\bottomrule
			\end{tabular}
		\caption{Summary of datasets used to evaluate and compare the proposed method}
		\label{tab:datasets}
	\end{table}

	\subsection{Experimental Framework}
	\label{sec:experimental-framework}
	To evaluate the proposed method, the oversamplers, metrics, datasets, and classifiers discussed in this section are used. Results are obtained by repeating 5-fold cross-validation five times. For each dataset, every metric is computed by averaging their values across runs. In addition to the arithmetic mean, the standard deviation is calculated.

	To achieve optimal results for all classifiers and oversamplers, a grid search procedure is used. For this purpose, each classifier and each oversampler specifies a set of possible values for every hyperparameter. Subsequently, all possible combinations of an algorithm's hyperparameters are generated and the algorithm is executed once for each combination. All metrics are used to score all resulting classifications, and the best value obtained for each metric is saved.

	To illustrate this process, consider an example with one oversampler and one classifier. The following list shows each algorithm with several combinations for each parameter.
	\begin{itemize}
		\item \ac{SMOTE}
		\begin{itemize}
			\item $knn$: 3,6
		\end{itemize}
		\item \ac{GBM}
		\begin{itemize}
			\item numberOfTrees: 10, 50
		\end{itemize}
	\end{itemize}
	The oversampling method \ac{SMOTE} is run two times, generating two different sets of training data. For each set of training data, the classifier \ac{LR} is executed twice. Therefore, the classifier will be executed four times. The possible combinations are:
	\begin{itemize}
		\item $knn = 3, numberOfTrees = 10$
		\item $knn = 3, numberOfTrees = 50$
		\item $knn = 6, numberOfTrees = 10$
		\item $knn = 6, numberOfTrees = 50$
	\end{itemize}
	If two metrics are used for scoring, both metrics score all four runs; the best result is saved. Note that one metric may find that the combination $knn = 3, numberOfTrees = 10$ is best, while a different combination might be best considering another metric. In this way, each oversampler and classifier combination is given the chance to optimize for each metric.
    
\section{Experimental Results}
\label{sec:experimental-results}

\begin{figure}[!htb]
\centering
\includegraphics[width=0.8\textwidth]{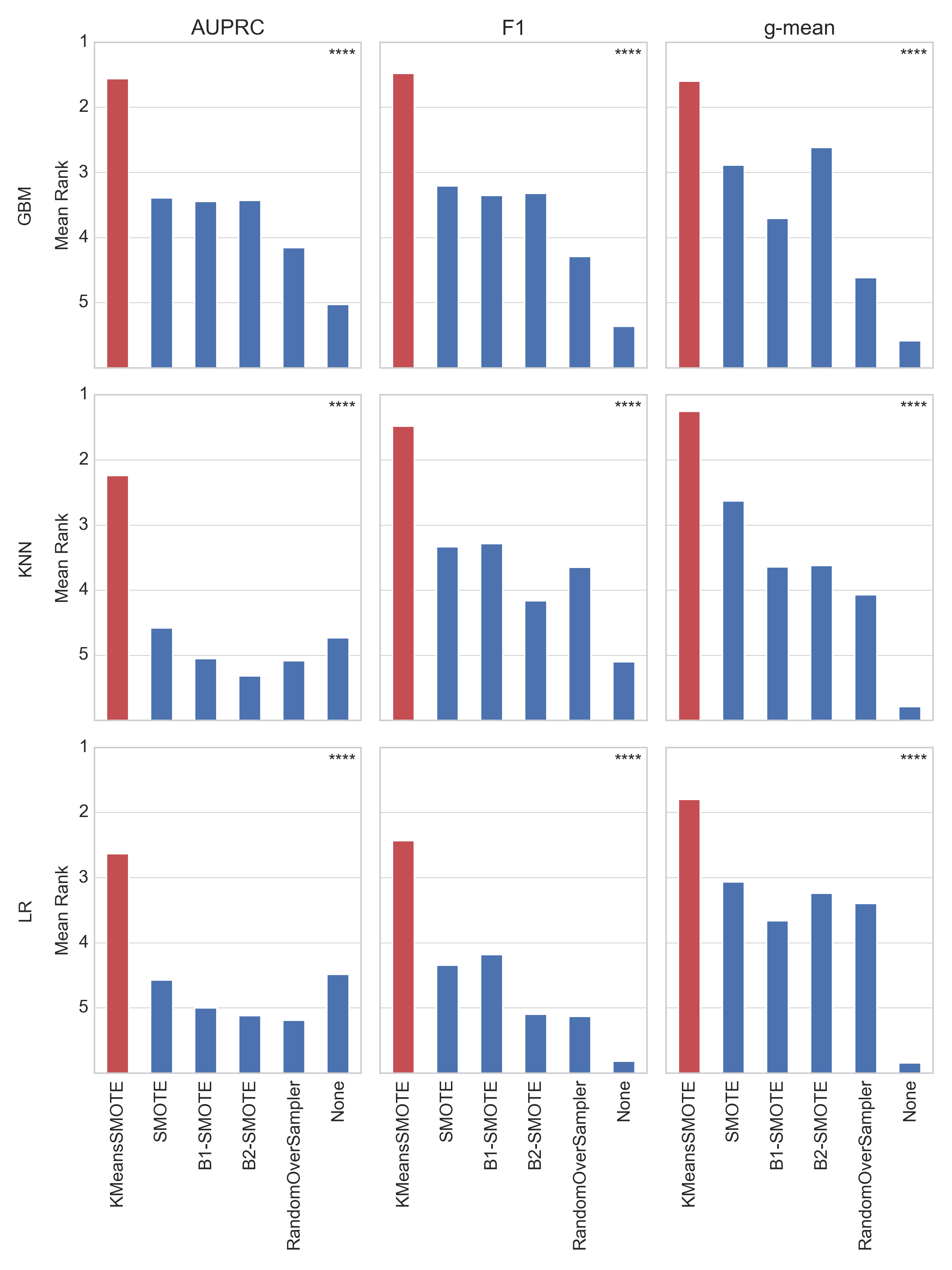}
\caption{Mean ranking of evaluated oversamplers for different classifiers and metrics}
\label{fig:ranking}
\end{figure}

In order to conclude whether any of the evaluated oversamplers perform consistently better than others, the cross-validated scores are used to derive a rank order, assigning rank one to the best performing and rank six to the worst performing technique~\citep{Demsar.2006}. This results in different rankings for each of five experiment repetitions, again partitioned by dataset, metric, and classifier. To aggregate the various rankings, each method's assigned rank is averaged across datasets and experiment repetitions. Consequently, a method's mean rank is a real number in the interval $[1.0,6.0]$. The mean ranking results for each combination of metric and classifier are shown in \cref{fig:ranking}.

By testing the null hypothesis that differences in terms of rank among oversamplers are merely a matter of chance, the Friedman~test~\citep{Friedman.1937} determines the statistical significance of the derived mean ranking. The test is chosen because it does not assume normality of the obtained scores~\citep{Demsar.2006}. At a significance level of $a = 0.05$, the null hypothesis is rejected for all evaluated classifiers and evaluation metrics. Therefore, the rankings are assumed to be significant.

The mean ranking shows that the proposed method outperforms other methods with regard to all evaluation metrics. Notably, the technique's superiority can be observed independently of the classifier. In six out of nine cases, k-means \ac{SMOTE} achieves a mean rank better than two, whereas in the other three cases the mean rank is at least three. Furthermore, k-means \ac{SMOTE} is the only technique with a mean ranking better than three with respect to F1 score and \ac{AUPRC}. This observation suggests that k-means \ac{SMOTE} improves classification results even when other oversamplers accomplish a similar rank as classifying without oversampling.

Generally, it can be observed that - aside from the proposed method - \ac{SMOTE}, borderline-\ac{SMOTE}1, and borderline-\ac{SMOTE}2 typically achieve the best results, while not oversampling usually earns the worst rank. Remarkably, \ac{LR} achieves a similar rank without oversampling as with \ac{SMOTE} with regard to \ac{AUPRC}, while both are only dominated by k-means \ac{SMOTE}. This indicates that the proposed method may improve classification results even when \ac{SMOTE} is not able to achieve any improvement versus the original training data.

\begin{figure}[ht]
	\centering
	\begin{minipage}{.48\textwidth}
		\centering
		\includegraphics[width=1\linewidth]{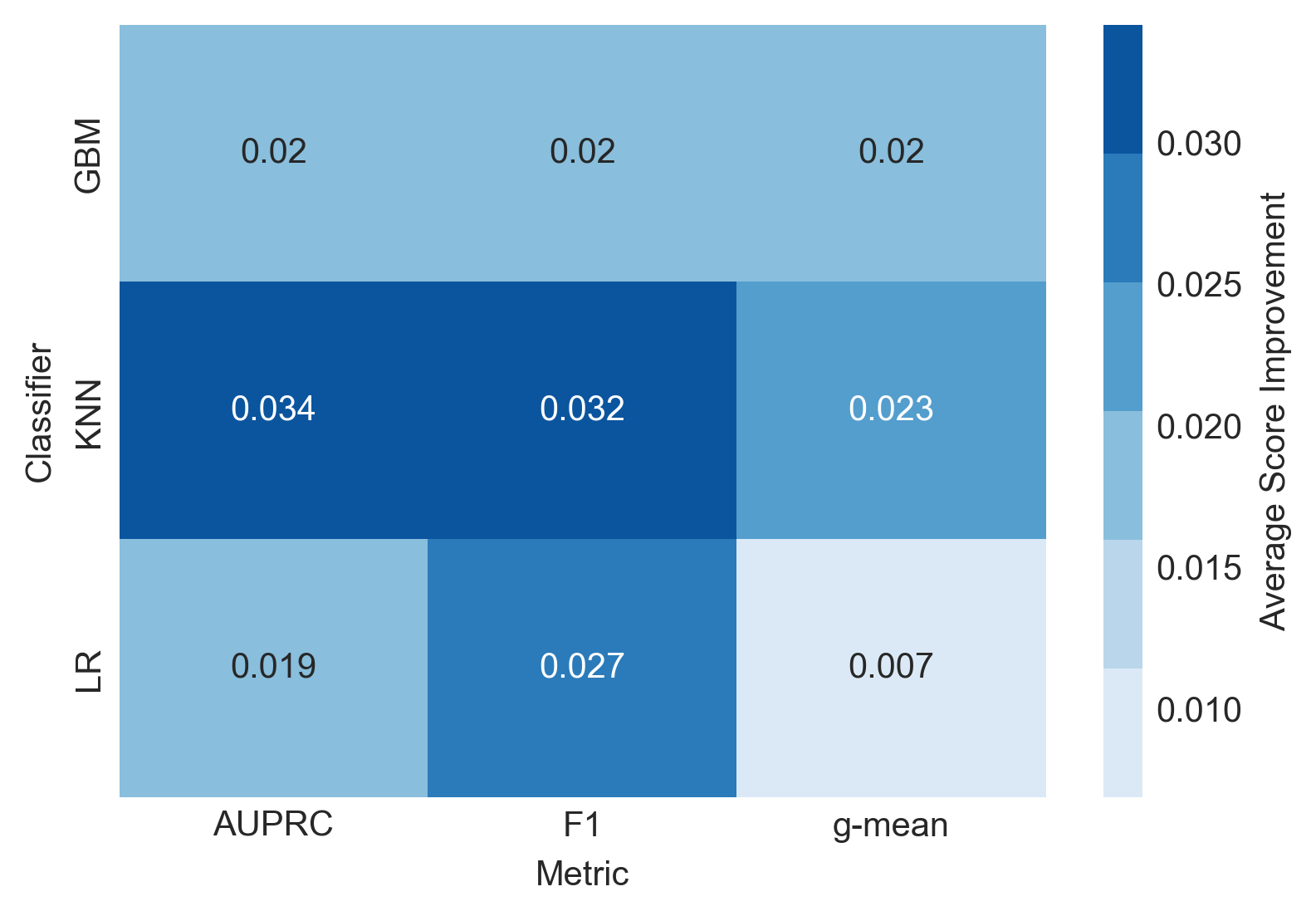}
		\captionof{figure}{Mean score improvement of the proposed method versus \acs{SMOTE} across datasets}
		\label{fig:avg_gains}
	\end{minipage}%
	\hfill
	\begin{minipage}{.48\textwidth}
		\centering
		\includegraphics[width=1\linewidth]{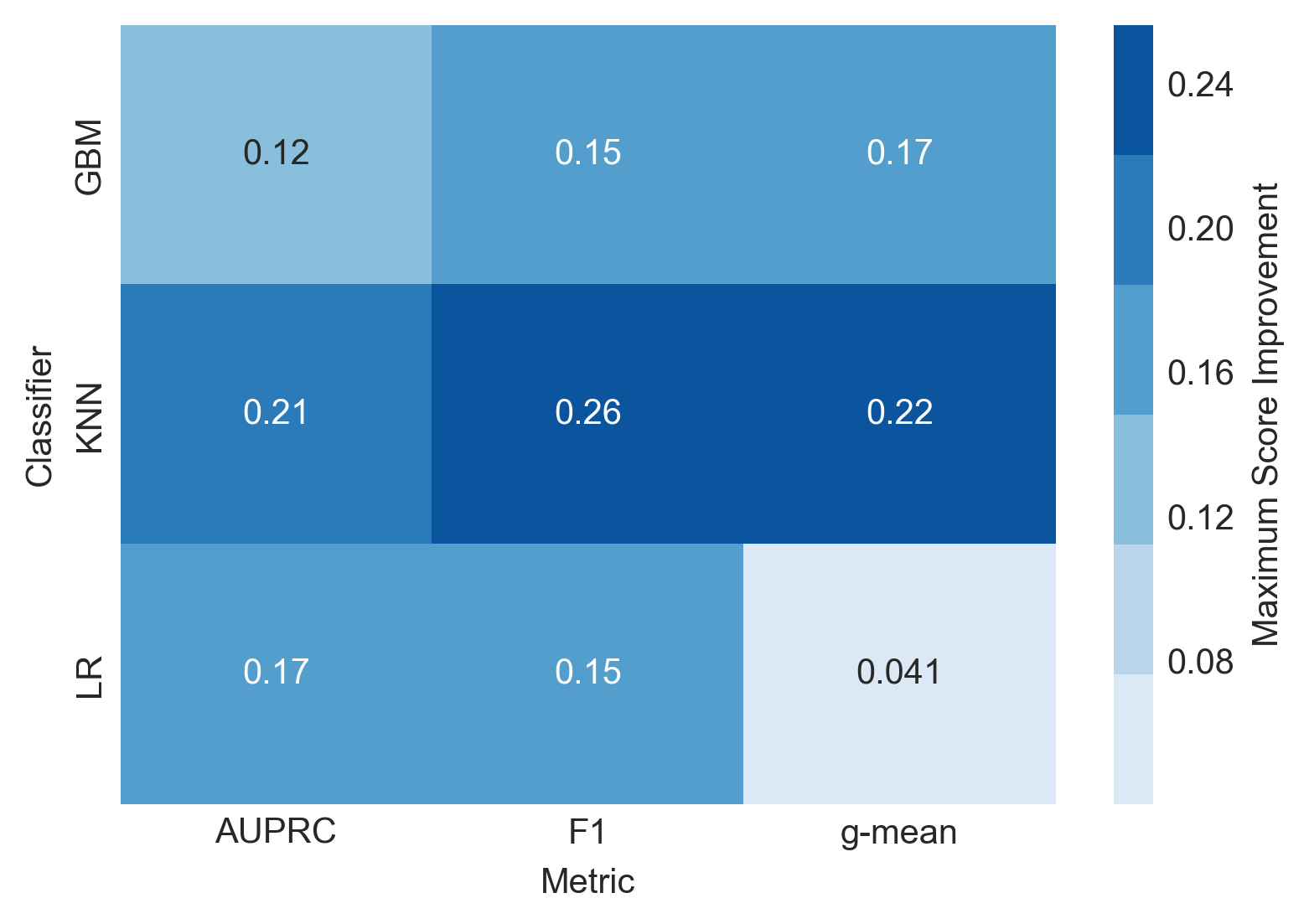}
		\captionof{figure}{Maximum score improvement of the proposed method versus \acs{SMOTE}}
		\label{fig:max_gains}
	\end{minipage}
\end{figure}

For a direct comparison to the baseline method, \ac{SMOTE}, the average optimal scores attained by k-means \ac{SMOTE} for each dataset are subtracted by the respective scores reached by \ac{SMOTE}. The resulting score improvements achieved by the proposed method are summarized in \cref{fig:avg_gains,,fig:max_gains}.

The \ac{KNN} classifier appears to profit most from the application of k-means \ac{SMOTE}, where maximum score improvements of more than 0.2 are observed across all metrics. The biggest mean score improvements are also achieved using \ac{KNN}, with a 0.034 average improvement with regard to the \ac{AUPRC} metric. It can further be observed that all classifiers benefit from the application of k-means \ac{SMOTE}. With one exception, maximum score improvements of more than 0.1 are achieved for all classifiers and metrics.

\begin{figure}[ht]
	\centering
	\includegraphics[width=0.87\textwidth]{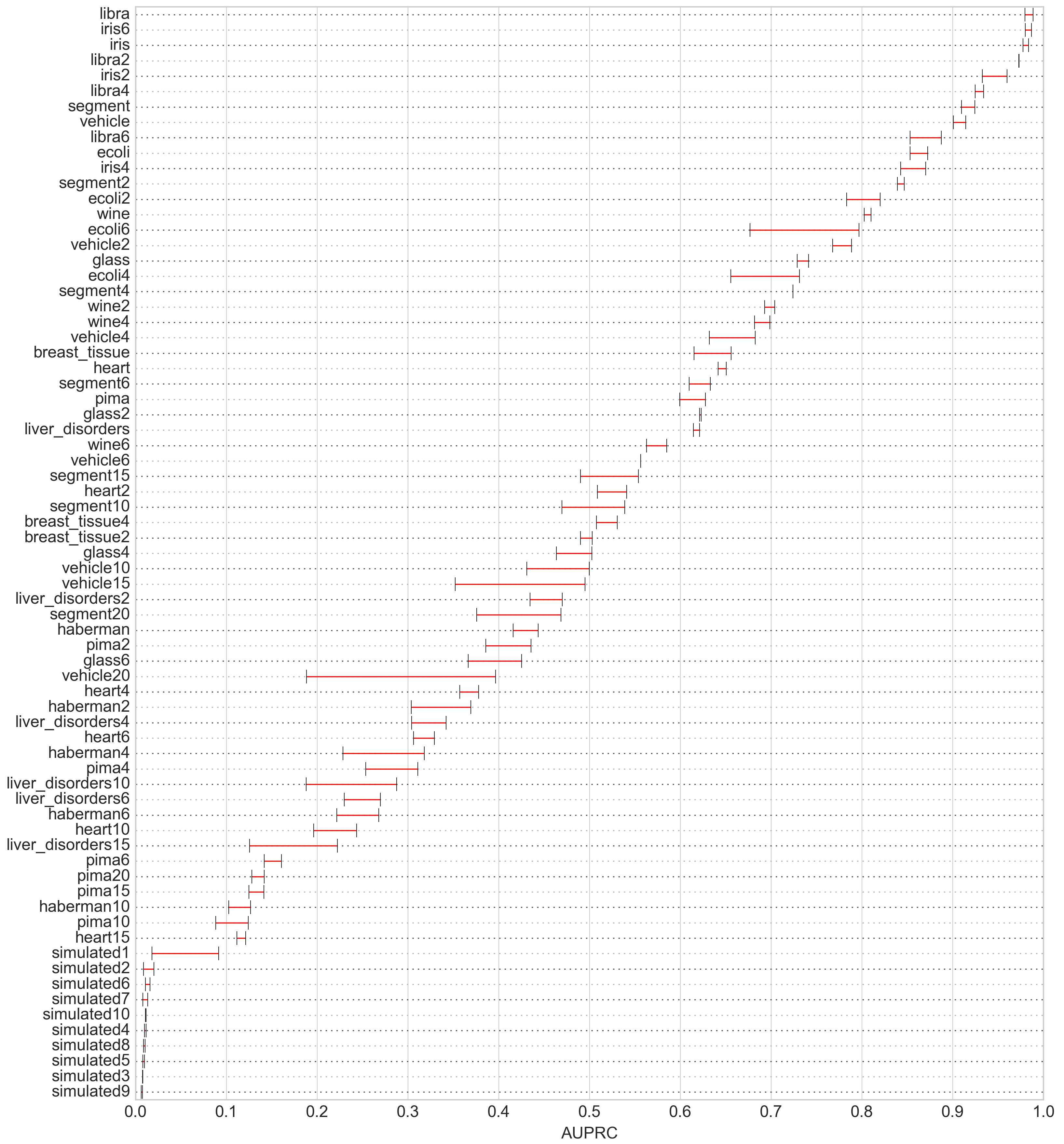}
	\caption{Performance of \acs{KNN} classifier trained on data oversampled with \acs{SMOTE} (left) and k-means \acs{SMOTE} (right)}
	\label{fig:comparison}
\end{figure}

Taking a closer look at the combination of classifier and metric which, on average, benefit most from the application of the proposed method, \cref{fig:comparison} shows the \ac{AUPRC} achieved by the two methods in conjunction with the \ac{KNN} classifier for each dataset. Although absolute scores and score differences between the two methods are dependent on the choice of metric and classifier, the general trend shown in the figure is observed for all other metrics and classifiers, which are omitted for clarity.

In the large majority of cases, k-means \ac{SMOTE} outperforms \ac{SMOTE}, proving the relevance of the clustering procedure. Only in 2 out of 71 datasets tested there were no improvements through the use of k-means \ac{SMOTE}. On average, k-means \ac{SMOTE} achieves an \ac{AUPRC} improvement of 0.034. The biggest gains of the proposed method appear to be occurring in the score range of 0.2 to 0.8. On the lower end of that range, k-means \ac{SMOTE} achieves an average gain of more than 0.2 compared to \ac{SMOTE} in the ``glass6'' dataset. Prediction of dataset ``vehicle15'' is improved from an \ac{AUPRC} of approximately 0.35 to 0.5. The third biggest gain occurs in the ``ecoli6'' dataset, where the proposed method obtains a score of 0.8 compared to less than 0.7 accomplished by \ac{SMOTE}. The score difference among oversamplers is smaller at the extreme ends of the scale. For nine of the simulated datasets, \ac{KNN} attains a score very close to zero independently of the choice of the oversampler. Similarly, for the datasets where an \ac{AUPRC} around 0.95 is attained (``libra'', ``libra2'', ``iris'', ``iris6''), gains of k-means \ac{SMOTE} are less than 0.02.

The discussed results are based on 5-fold cross-validation with five repetitions, using tests to assure statistical significance. Mean ranking results show that oversampling using k-means \ac{SMOTE} improves the performance of different evaluated classifiers on imbalanced data. In addition, the proposed oversampler dominates all evaluated oversampling methods in mean rankings regardless of the classifier. Studying absolute gains of the proposed algorithm compared to the baseline method, it is found that all classifiers used in the study benefit from the clustering and sample distribution procedure of k-mean \ac{SMOTE}. Additionally, almost all datasets profit from the proposed method. Generally, classification problems which are neither very easy nor very difficult profit most, allowing significant score increases of up to 0.26. It is therefore concluded that k-means \ac{SMOTE} is effective in generating samples which aid classifiers in the presence of imbalance.

\section{Conclusion}
\label{sec:conclusion}
Imbalanced data poses a difficult task for many classification algorithms. Resampling training data toward a more balanced distribution is an effective way to combat this issue independently of the choice of the classifier. However, balancing classes by merely duplicating minority class instances encourages overfitting, which in turn degrades the model's performance on unseen data. Techniques which generate artificial samples, on the other hand, often suffer from a tendency to generate noisy samples, impeding the inference of class boundaries. Moreover, most existing oversamplers do not counteract imbalances within the minority class, which is often a major issue when classifying class-imbalanced datasets. For oversampling to effectively aid classifiers, the amplification of noise should be avoided by detecting safe areas of the input space where class regions do not overlap. Additionally, any imbalance within the minority class should be identified and samples are to be generated as to level the distribution.

The proposed method achieves these properties by clustering the data using k-means, allowing to focus data generation on crucial areas of the input space. A high ratio of minority observations is used as an indicator that a cluster is a safe area. Oversampling only safe clusters enables k-means \ac{SMOTE} to avoid noise generation. Furthermore, the average distance among a cluster's minority samples is used to discover sparse areas. Sparse minority clusters are assigned more synthetic samples, which alleviates within-class imbalance. Finally, overfitting is discouraged by generating genuinely new observations using \ac{SMOTE} rather than replicating existing ones.

Empirical results show that training various types of classifiers using data oversampled with k-means \ac{SMOTE} leads to better classification results than training with unmodified, imbalanced data. More importantly, the proposed method consistently outperforms the most widely available oversampling techniques such as \ac{SMOTE}, borderline-\ac{SMOTE}, and random oversampling. The biggest gains appear to be achieved in classification problems which are neither extremely difficult nor extremely simple. The results are statistically robust and apply to various metrics suited for the evaluation of imbalanced data classification.

The effectiveness of the algorithm is accomplished without high complexity. The method's components, k-means clustering and \ac{SMOTE} oversampling, are simple and readily available in many programming languages, so that practitioners and researchers may easily implement and use the proposed method in their preferred environment. Further facilitating practical use, an implementation of k-means \ac{SMOTE} in the python programming language is made available (see \url{https://github.com/felix-last/kmeans_smote}) based on the imbalanced-learn framework \citep{Lemaitre.2017}.

A prevalent issue in classification tasks, data imbalance is exhibited naturally in many important real-world applications. As the proposed oversampler can be applied to rebalance any dataset and independently of the chosen classifier, its potential impact is substantial. Among others, k-means \ac{SMOTE} may, therefore, contribute to the prevention of credit card fraud, the diagnosis of diseases, as well as the detection of abnormalities in environmental observations.

Future work may consequently focus on applying k-means \ac{SMOTE} to various other real-world problems. Additionally, finding optimal values of $k$ and other hyperparameters is yet to be guided by rules of thumb, which could be deducted from further analyses of the relationship between optimal hyperparameters for a given dataset and the dataset's properties.

\bibliographystyle{apalike}
\bibliography{references}

\begin{thebibliography}{}

\bibitem[Batista et~al., 2004]{Batista.2004}
Batista, G. E. A. P.~A., Prati, R.~C., and Monard, M.~C. (2004).
\newblock A study of the behavior of several methods for balancing machine
  learning training data.
\newblock {\em ACM SIGKDD Explorations Newsletter}, 6(1):20--29.

\bibitem[Bunkhumpornpat et~al., 2009]{Bunkhumpornpat.2009}
Bunkhumpornpat, C., Sinapiromsaran, K., and Lursinsap, C. (2009).
\newblock Safe-level-smote: Safe-level-synthetic minority over-sampling
  technique for handling the class imbalanced problem.
\newblock In {\em Lecture Notes in Computer Science (including subseries
  Lecture Notes in Artificial Intelligence and Lecture Notes in
  Bioinformatics)}, volume 5476 LNAI, pages 475--482.

\bibitem[Chawla et~al., 2002]{Chawla.2002}
Chawla, N.~V., Bowyer, K.~W., Hall, L.~O., and Kegelmeyer, W.~P. (2002).
\newblock Smote: Synthetic minority over-sampling technique.
\newblock {\em Journal of Artificial Intelligence Research}, 16:321--357.

\bibitem[Chawla et~al., 2004]{Chawla.2004}
Chawla, N.~V., Japkowicz, N., and Drive, P. (2004).
\newblock Editorial: Special issue on learning from imbalanced data sets.
\newblock {\em ACM SIGKDD Explorations Newsletter}, 6(1):1--6.

\bibitem[Cieslak et~al., 2006]{Cieslak.2006b}
Cieslak, D.~A., Chawla, N.~V., and Striegel, A. (2006).
\newblock Combating imbalance in network intrusion datasets.
\newblock In {\em Granular Computing, 2006 IEEE International Conference on},
  pages 732--737. IEEE.

\bibitem[Demšar, 2006]{Demsar.2006}
Demšar, J. (2006).
\newblock Statistical comparisons of classifiers over multiple data sets.
\newblock {\em Journal of Machine learning research}, 7(Jan):1--30.

\bibitem[Domingos, 1999]{Domingos.1999}
Domingos, P. (1999).
\newblock Metacost: A general method for making classifiers.
\newblock In {\em Proceedings of the 5th International Conference on Knowledge
  Discovery and Data Mining}, pages 155--164.

\bibitem[Douzas and Bacao, 2017]{Douzas.2017}
Douzas, G. and Bacao, F. (2017).
\newblock Self-organizing map oversampling (somo) for imbalanced data set
  learning.
\newblock {\em Expert Systems with Applications}, 82:40--52.

\bibitem[Estabrooks et~al., 2004]{Estabrooks.2004}
Estabrooks, A., Jo, T., and Japkowicz, N. (2004).
\newblock A multiple resampling method for learning from imbalanced data sets.
\newblock {\em Computational intelligence}, 20(1):18--36.

\bibitem[Fern{\'a}ndez et~al., 2013]{Fernandez.2013}
Fern{\'a}ndez, A., L{\'o}pez, V., Galar, M., {Del Jesus}, M.~J., and Herrera,
  F. (2013).
\newblock Analysing the classification of imbalanced data-sets with multiple
  classes: Binarization techniques and ad-hoc approaches.
\newblock {\em Knowledge-Based Systems}, 42:97--110.

\bibitem[Fix and {Hodges Jr.}, 1951]{Fix.1951}
Fix, E. and {Hodges Jr.}, J. (1951).
\newblock Discriminatory analysis - nonparametric discrimination: Consistency
  properties.

\bibitem[Friedman, 2001]{Friedman.2001}
Friedman, J.~H. (2001).
\newblock Greedy function approximation: A gradient boosting machine.
\newblock {\em Annals of Statistics}, 29(5):1189--1232.

\bibitem[Friedman, 1937]{Friedman.1937}
Friedman, M. (1937).
\newblock The use of ranks to avoid the assumption of normality implicit in the
  analysis of variance.
\newblock {\em Journal of the American Statistical Association}, 32(200):675.

\bibitem[Galar et~al., 2012]{Galar.2012}
Galar, M., Fernandez, A., Barrenechea, E., Bustince, H., and Herrera, F.
  (2012).
\newblock A review on ensembles for the class imbalance problem: Bagging-,
  boosting-, and hybrid-based approaches.

\bibitem[Guyon, 2003]{Guyon.2003}
Guyon, I. (2003).
\newblock Design of experiments of the nips 2003 variable selection benchmark.

\bibitem[Han et~al., 2005]{Han.2005}
Han, H., Wang, W.-Y., and Mao, B.-H. (2005).
\newblock Borderline-smote: A new over-sampling method in imbalanced data sets
  learning.
\newblock {\em Advances in intelligent computing}, 17(12):878--887.

\bibitem[Hawkins, 2004]{Hawkins.2004}
Hawkins, D.~M. (2004).
\newblock The problem of overfitting.
\newblock {\em Journal of chemical information and computer sciences},
  44(1):1--12.

\bibitem[He and Garcia, 2009]{He.2009}
He, H. and Garcia, E.~A. (2009).
\newblock Learning from imbalanced data.
\newblock {\em IEEE Transactions on Knowledge and Data Engineering},
  21(9):1263--1284.

\bibitem[Holte et~al., 1989]{Holte.1989}
Holte, R.~C., Acker, L., Porter, B.~W., et~al. (1989).
\newblock Concept learning and the problem of small disjuncts.
\newblock In {\em IJCAI}, volume~89, pages 813--818.

\bibitem[Japkowicz, 2013]{Japkowicz.2013}
Japkowicz, N. (2013).
\newblock Assessment metrics for imbalanced learning.
\newblock In He, H. and Ma, Y., editors, {\em Imbalanced learning}, pages
  187--206. {John Wiley {\&} Sons}.

\bibitem[Jo and Japkowicz, 2004]{Jo.2004}
Jo, T. and Japkowicz, N. (2004).
\newblock Class imbalances versus small disjuncts.
\newblock {\em ACM SIGKDD Explorations Newsletter}, 6(1):40--49.

\bibitem[Kotsiantis et~al., 2006]{Kotsiantis.2006}
Kotsiantis, S., Kanellopoulos, D., and Pintelas, P. (2006).
\newblock Handling imbalanced datasets: A review.
\newblock {\em Science}, 30(1):25--36.

\bibitem[Kotsiantis et~al., 2007]{Kotsiantis.2007}
Kotsiantis, S., Pintelas, P., Anyfantis, D., and Karagiannopoulos, M. (2007).
\newblock Robustness of learning techniques in handling class noise in
  imbalanced datasets.

\bibitem[Lemaître et~al., 2017]{Lemaitre.2017}
Lemaître, G., Nogueira, F., and Aridas, C.~K. (2017).
\newblock Imbalanced-learn: {A} {Python} {Toolbox} to {Tackle} the {Curse} of
  {Imbalanced} {Datasets} in {Machine} {Learning}.
\newblock {\em Journal of Machine Learning Research}, 18(17):1--5.

\bibitem[Lichman, 2013]{Lichman.2013}
Lichman, M. (2013).
\newblock Uci machine learning repository.

\bibitem[Ma and Fan, 2017]{Ma.2017}
Ma, L. and Fan, S. (2017).
\newblock Cure-smote algorithm and hybrid algorithm for feature selection and
  parameter optimization based on random forests.
\newblock {\em BMC bioinformatics}, 18(1):169.

\bibitem[MacQueen, 1967]{MacQueen.1967}
MacQueen, J. (1967).
\newblock Some methods for classification and analysis of multivariate
  observations.
\newblock In {\em Proceedings of the fifth Berkeley symposium on mathematical
  statistics and probability}, volume~1, pages 281--297.

\bibitem[McCullagh, 1984]{McCullagh.1984}
McCullagh, P. (1984).
\newblock Generalized linear models.
\newblock {\em European Journal of Operational Research}, 16(3):285--292.

\bibitem[Nekooeimehr and Lai-Yuen, 2016]{Nekooeimehr.2016}
Nekooeimehr, I. and Lai-Yuen, S.~K. (2016).
\newblock Adaptive semi-unsupervised weighted oversampling (a-suwo) for
  imbalanced datasets.
\newblock {\em Expert Systems with Applications}, 46:405--416.

\bibitem[Nickerson et~al., 2001]{Nickerson.2001}
Nickerson, A., Japkowicz, N., and Milios, E.~E. (2001).
\newblock Using unsupervised learning to guide resampling in imbalanced data
  sets.
\newblock In {\em AISTATS}.

\bibitem[Pedregosa et~al., 2011]{Pedregosa.2011}
Pedregosa, F., Varoquaux, G., Gramfort, A., Michel, V., Thirion, B., Grisel,
  O., Blondel, M., Prettenhofer, P., Weiss, R., Dubourg, V., Vanderplas, J.,
  Passos, A., Cournapeau, D., Brucher, M., Perrot, M., and Duchesnay, E.
  (2011).
\newblock Scikit-learn: Machine learning in python.
\newblock {\em Journal of Machine learning research}, 12:2825--2830.

\bibitem[Prati et~al., 2004]{Prati.2004}
Prati, R.~C., Batista, G., and Monard, M.~C. (2004).
\newblock Learning with class skews and small disjuncts.
\newblock In {\em SBIA}, pages 296--306.

\bibitem[Provost, 2000]{Provost.2000}
Provost, F. (2000).
\newblock Machine learning from imbalanced data sets 101.
\newblock In {\em Proceedings of the AAAI'2000 workshop on imbalanced data
  sets}, volume~68, pages 1--3. AAAI Press.

\bibitem[Santos et~al., 2015]{Santos.2015}
Santos, M.~S., Abreu, P.~H., Garc{\'i}a-Laencina, P.~J., Sim{\~a}o, A., and
  Carvalho, A. (2015).
\newblock A new cluster-based oversampling method for improving survival
  prediction of hepatocellular carcinoma patients.
\newblock {\em Journal of biomedical informatics}, 58:49--59.

\bibitem[Sculley, 2010]{Sculley.2010}
Sculley, D. (2010).
\newblock Web-scale k-means clustering.
\newblock In {\em Proceedings of the 19th international conference on World
  wide web}, pages 1177--1178, 1772690. ACM.

\bibitem[Song et~al., 2016]{Song.2016}
Song, J., Huang, X., Qin, S., and Song, Q. (2016).
\newblock A bi-directional sampling based on k-means method for imbalance text
  classification.
\newblock In {\em Computer and Information Science (ICIS), 2016 IEEE/ACIS 15th
  International Conference on}, pages 1--5.

\bibitem[Weiss et~al., 2007]{Weiss.2007}
Weiss, G.~M., McCarthy, K., and Zabar, B. (2007).
\newblock Cost-sensitive learning vs. sampling: Which is best for handling
  unbalanced classes with unequal error costs?
\newblock {\em DMIN}, 7:35--41.

\end{thebibliography}

\end{document}